\documentclass[11pt]{article}

\usepackage[final]{acl}

\usepackage{times}
\usepackage{latexsym}
\usepackage[T1]{fontenc}
\usepackage[utf8]{inputenc}
\usepackage{microtype}
\usepackage{inconsolata}
\usepackage{graphicx}
\usepackage{booktabs}
\usepackage{multirow}
\usepackage{amsmath}
\usepackage{amssymb}
\usepackage{hyperref}
\usepackage{tabularx}
\usepackage{seqsplit}
\usepackage{color}
\usepackage[table]{xcolor}

\title{ReflectRec: Cycle-Consistency Preference Verification\\
for LLM-based Recommendation}

\author{
  Dae Kyoo Kim$^{1,2}$ \quad 
  Heejung Cho$^{1,3,*}$ \quad 
  Sungmoo Hong$^{1,4}$ \quad 
  Eunil Park$^{1,5,}$\thanks{Corresponding authors.} \\
  $^1$Sungkyunkwan University  \quad
  $^2$NAVER Corp.  \quad
  $^3$KIAT  \quad
  $^4$KT Corp. \quad
  $^5$Jaume I University \\
  \texttt{\{kdg7447, heejungcho, isean, eunilpark\}@skku.edu}
}

\begin{document}
\maketitle

\begin{abstract}
CoT-style LLM recommenders generate a natural-language preference note as a chain-of-thought prefix to ranking, but the note is trusted without verification, and the ranker never consults behaviorally similar users. We propose \textbf{ReflectRec}, organized around a preference-to-history round-trip verifier whose recall score serves as an operational signal of preference-to-history recoverability, enabling conditional regeneration for low-recoverability preferences and selection among regeneration attempts. Around this verifier, we wire a semantic Stage-1 prompt, conditional regeneration, and Forward-Look retrieval that injects future watches of behaviorally similar users as a collaborative hint. On MovieLens-100K and MovieLens-1M, the full pipeline outperforms CoT4Rec-ICL by $+16.25$ / $+4.46$\,pp Hit@10, with positive gains across all ValScore buckets and three LLM backbones (Qwen3-30B, Qwen2.5-72B, and Llama-3.3-70B). We also observe a similar verifier-guided regeneration pattern on Amazon Electronics.
\end{abstract}

\section{Introduction}
\label{sec:intro}
Sequential and conversational recommendation is rapidly being reshaped by large language models~\citep{geng2022p5,hou2023zeroshot,zhang2023collm,wu2024survey}. Unlike classical collaborative filtering, which compresses interactions into latent vectors~\citep{koren2009mf,rendle2009bpr}, Large Language Model(LLM) based recommenders represent histories and items in natural language and can reason about \emph{why} a user may like an item, producing human-readable rationales~\citep{wei2022cot,bao2023tallrec,gao2023chatrec,wang2025rallrecplus}.

A representative instance is \textsc{CoT4Rec}~\citep{cot4rec2025}, which factors ranking into two stages: a \emph{preference analyzer} turns a raw history into a short natural language preference note, and a \emph{recommender} consumes the history, preference, and candidate list to predict the next item. Such pipelines inherit the interpretability of language-based recommendations while delivering strong empirical gains.

However, two weaknesses persist in Chain of Thought (CoT) style pipelines. The first is the \emph{absence of verification for intermediate preferences.} LLM outputs are sensitive to prompt variation~\citep{lu2022fantastically}, hallucinate when forced to externalize user preferences~\citep{chen2025preferences}, and can hurt downstream rankers when they are too abstract or factually inconsistent. However, no established mechanism measures the per-sample faithfulness of the preference or repairs low-quality preferences before they reach the ranker. The second is the \emph{absence of explicit reference to behaviorally similar histories from other users}. At inference, the ranker attends only to one user's history and its compressed preference; it never consults what users with similar local viewing patterns recently consumed. The local collaborative cue that classical recommenders draw heavily on is missing, and the \emph{lost-in-the-middle} phenomenon~\citep{liu2024lostmiddle} aggravates the problem for LLMs, whose attention over long histories is uneven.

\begin{figure*}[t]
\centering
\includegraphics[width=\linewidth]{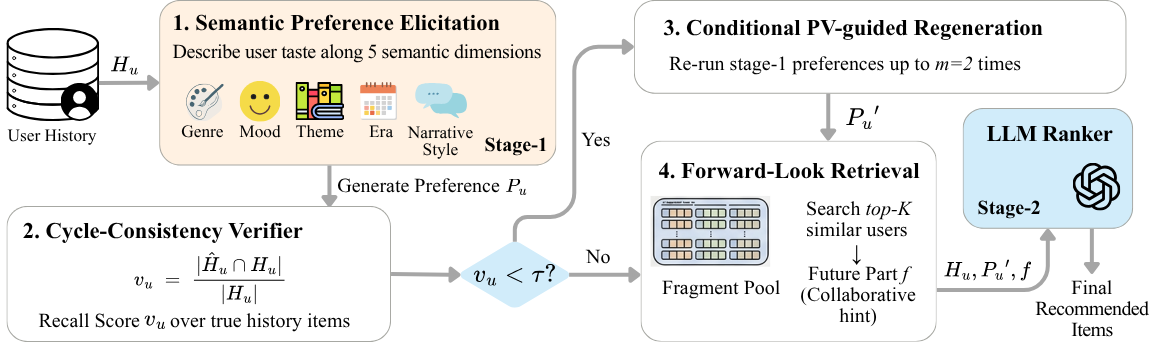}
\caption{\textsc{ReflectRec} overview. Stage-1 elicits a structured user preference $P_u$; the cycle-consistency verifier yields a score of preference-to-history recoverability, which gates regeneration, selects the regeneration attempt with the highest recoverability, and is used to analyze downstream module effects. Conditional regeneration regenerates low-recoverability preferences, and Forward-Look retrieval injects the \emph{future watches} of behaviorally similar users into the ranker prompt.\label{fig:framework}}
\end{figure*}

We introduce \textbf{ReflectRec} (Fig.~\ref{fig:framework}), a recommendation pipeline built around a single unifying signal: the \emph{cycle-consistency verification score}. This score is computed by feeding the generated preference $P_u$ back into the LLM alongside 50 candidates, the user's true history items plus random negatives, and measuring the recall $v_u$ over true history items. This single value is used for four functions: (i) an operational signal of preference-to-history recoverability, (ii) \emph{gating} of regeneration, (iii) \emph{selection} of the regeneration attempt with the highest recoverability score, and (iv) \emph{diagnosis} of how downstream module effects vary across ValScore levels. Around this verifier, we wire four modules, which are described below.

\paragraph{Semantic Preference Elicitation.} A single LLM call instructs the analyzer to describe user taste along five semantic dimensions, genre, mood, theme, era, and narrative style, without quoting any movie titles from the user's history. No embedding or retrieval is involved at this stage; the resulting preference note serves as the input substrate for all three downstream modules (verifier, regeneration, and forward-looking).

\paragraph{Cycle-Consistency Verifier (four roles).} It produces the recall $v_u$ defined above and uses it simultaneously for measurement, gating, selection, and diagnosis.

\paragraph{Conditional PV-Guided Regeneration.} Only samples with $\mathrm{ValScore} < \tau{=}0.4$ (the \textsc{low} bucket) have their preference re-run up to $m{=}2$ times. The verifier scores each attempt and retains the highest score.

\paragraph{Forward-Look Retrieval.} From a fragment pool built on the training split only, each fragment is a tuple $(h, f)$ where $h$ is a 5 item history window and $f$ is the user's \emph{next 5 items actually watched}, we retrieve top-$k$ fragments by history similarity and inject the \emph{future} part $\{f\}$ into the ranker prompt as a collaborative hint. Empirically, this retrieval contributes most to the \textsc{mid}/\textsc{high}/\textsc{pass} buckets (\S\ref{sec:bucket}).

\paragraph{Empirical Findings.}
We evaluate on MovieLens-100K ($n{=}843$) and MovieLens-1M ($n{=}1500$) with three LLM backbones (Qwen3-30B, Qwen2.5-72B, and Llama-3.3-70B). We examine whether the cycle-consistency verifier provides a per-sample operational signal of preference-to-history recoverability; conditional regeneration improves Hit@K for low-recoverability samples; Forward-Look retrieval provides a collaborative signal complementary to regeneration; and each module contributes to overall performance with effects that vary across ValScore buckets.

The full pipeline reaches Hit@10 $=31.67\%$ on ML-100K and $39.93\%$ on ML-1M. Both gains are statistically significant ($p{=}0.014$, $p{=}0.036$, paired bootstrap). In the random-future control, similarity-based Forward-Look yields $+3.32$ / $+3.14$\,pp higher Hit@10 than arbitrary future hints. In the unified pool ($n{=}2343$), all four ValScore buckets exhibit positive $\Delta$Hit@10.

\paragraph{Contributions.}
We introduce a task-grounded \emph{cycle-consistency verifier} that checks whether an LLM-generated preference note can recover the user's history items from a candidate pool, with a single recall score simultaneously serving four operational roles: measure, gate, select, and diagnose. Around this verifier, we build \textbf{ReflectRec}, a four-module pipeline that yields complementary gains across ValScore buckets, supported by a random-future control, paired bootstrap CIs, and consistent results across three LLM backbones.

\section{Related Work\label{sec:related}}
\paragraph{LLM as a recommender.}
Recent advancements in language modeling shifted personalized recommendations from traditional numerical prediction~\citep{koren2009mf,rendle2009bpr} to linguistic reasoning by reformulating them as a text-to-text generation task~\citep{geng2022p5}. Based on this generative paradigm, contemporary approaches have incorporated diverse strategies, such as zero-shot inference~\citep{hou2023zeroshot}, collaborative embedding fusion~\citep{zhang2023collm,liao2024llara}, data augmentation~\citep{wang2024cold}, and external knowledge grounding~\citep{xi2024openworld}.

\paragraph{Chain-of-thought recommendation.}
\citet{cot4rec2025} introduce the two-stage analyzer/ranker pipeline that we extend. \citet{ren2024rla} learn LLM-aligned representations; all assume that intermediate rationales help downstream recommendations, but none explicitly measure the rationale’s faithfulness to the user history.

\paragraph{Retrieval-augmented recommendation.}
RAG is the standard for knowledge-intensive NLP \cite{lewis2020rag}. \citet{wang2025rallrecplus} combine RAG with LLM recommendations by retrieving similar items. A natural extension is to retrieve behavior fragments (short interaction segments of other users); within that space, we make a specific design choice: the retrieval target should be aligned with the next-item prediction task rather than with the preference itself. Therefore, we do not score fragments by similarity to the preference; instead, we score them by similarity of their history part to the test user’s history and feed their future part (the next five items actually watched after the history window) into the ranker as a collaborative hint.

\paragraph{Quality and self-verification of LLM intermediates.}
Prompt sensitivity \cite{lu2022fantastically}, hallucinated preferences \cite{chen2025preferences}, and self-correction via backtracking \cite{sadeghi2025backtracking} all motivate the explicit verification of LLM intermediates. \citet{liu2024lostmiddle} show that LLM attention degrades in the middle of the long contexts. Round-trip/cycle-consistency has a long history in NLP and vision, back-translation for translation quality, QAGS and FactCC for summarization fact-checking, and self-consistency CoT for reasoning. An adjacent example is \citet{kim2025lost}, who use reverse reconstruction (recovering the original problem from a reasoning trajectory and answer) as a mutual-verification signal to improve the logical-semantic consistency of small language models on mathematical reasoning. We apply the same round-trip idea to preference verification in LLM-based recommendations, where a single verification score simultaneously drives the measure, gate, and select operations and diagnoses the subpopulation effects of downstream modules.

\section{Method\label{sec:method}}
\subsection{Problem Formulation}
Let $\mathcal{U}$ denote the set of users and $\mathcal{I}$ the set of items. Each user $u \in \mathcal{U}$ has a chronologically ordered history $H_u = [i_1, \ldots, i_T]$ with $i_t \in \mathcal{I}$. Given a candidate set $C_u \subset \mathcal{I}$, the recommender predicts the next item.

\begin{equation}
\begin{aligned}
P_u^{(0)} &= f_\text{pref}(H_u), \\
v_u &= f_\text{val}(H_u, P_u^{(0)}), \\
P_u^{'} &= f_\text{regen}\bigl(H_u, P_u^{(0)}, v_u\bigr), \\
R_u &= f_\text{ret}(H_u; \mathcal{F}_\text{pool}), \\
\hat{y}_u &= f_\text{rec}(H_u, P_u^{'}, R_u, C_u)
\end{aligned}
\label{eq:overall}
\end{equation}
$f_\text{pref}$ produces the semantic preference, $f_\text{val}$ the verification score, $f_\text{regen}$ returns the regenerated preference when $v_u < \tau$ and the original otherwise, $f_\text{ret}$ retrieves Forward-Look fragments, and $f_\text{rec}$ is the final ranker. All LLM-based stages use the same backbone.

\subsection{Stage 1: Semantic Preference Prompt\label{sec:stage1}}
\paragraph{Design Principle.}
The ranker prompt already contains raw history $H_u$. For the preference $P_u$ to contribute information the ranker cannot infer elsewhere, $P_u$ must be \emph{complementary} to $H_u$ rather than redundant with it: $P_u$ should encode descriptive content (genre, sub-genre, mood and emotional tone, thematic content, era or stylistic period, narrative style) rather than restating items.

\paragraph{Prompt.}
The Stage-1 prompt asks the LLM for a 60--100 word descriptive preference along the five axes above. It discourages verbatim title quotations; any title that still appears is detected by exact case-insensitive substring matching and is used as a tie-breaker when selecting among regeneration attempts (\S\ref{sec:regen}).

\subsection{Cycle-Consistency Preference Verification\label{sec:pv}}

\begin{figure*}[!t]
\centering
\includegraphics[width=\linewidth]{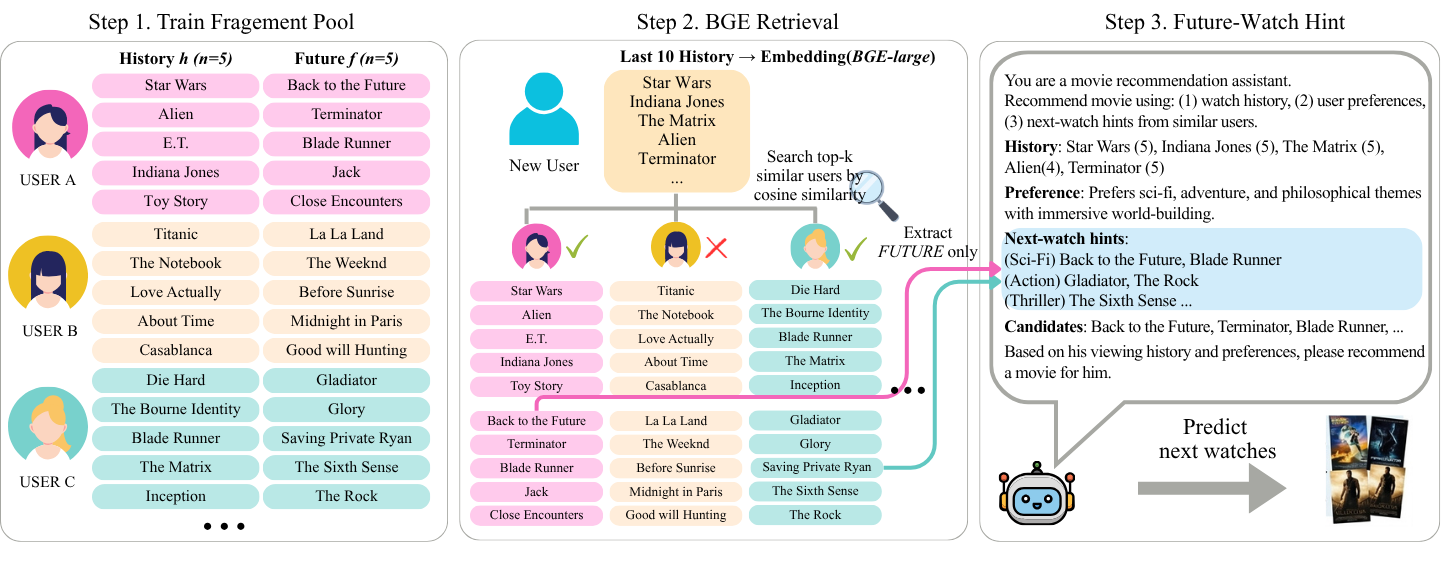}
\caption{Training users' histories are split into history $h$ and future $f$ fragments to build a fragment pool (Step 1). A new user's last 10 interactions are embedded via BGE-large and matched to the top-$K$ similar users by cosine similarity (Step 2). Only the future fragments of matched users are extracted and injected as next-watch hints into the LLM ranker prompt alongside the user's history and preferences (Step 3).}
\label{fig:validator}
\end{figure*}

\paragraph{Principle: preference $\to$ history round-trip.}
Directly evaluating how well a preference note $P_u$ reflects a user's preference is difficult. ``How well does this text describe the user's taste?'' has no direct ground truth, and LLM-judge-style self-rating is prone to prompt sensitivity and sycophancy~\citep{lu2022fantastically}. We measure how many items from the history $H_u$ that generated $P_u$ can be recovered from $P_u$. We use this reconstruction recall as an operational signal of preference-to-history recoverability.

\paragraph{Verification score.}
Given $P_u$, the verifier LLM receives a mixed pool of 50 candidates (the user's true history items plus random negatives) and is asked to select the ones that the user has most likely watched. The recall over true history items is:
\begin{equation}
\mathrm{ValScore}(P_u) \;=\; \frac{|\hat{H}_u \cap H_u|}{|H_u|}.
\end{equation}
A higher ValScore indicates that more of the user's observed history items can be recovered from the generated preference. We therefore use ValScore as a score of preference-to-history recoverability.

\paragraph{Round-trip versus LLM-judge baseline.}
An alternative design would ask the LLM to rate the preference text on a 1--5 scale. Such self-ratings tend to (i) give consistently high scores to abstract, well-written preferences (sycophancy) and (ii) be sensitive to minor wording changes in the rating prompt. In contrast, round-trip verification is grounded in \emph{concrete, observable output} (the recall over history items) and therefore naturally penalizes hallucinated abstract preferences, as they cannot supply enough cues to distinguish the true history items from random negatives.

\paragraph{Candidate pool of size 50.}
The pool of 50 contains the user's true history items and random negatives drawn from $\mathcal{I} \setminus H_u$. With an average $|H_u|=10$, the positive-to-negative ratio is 1:4. We use this 1:4 construction as a descriptive reference for interpreting ValScore. The regeneration operating threshold $\tau=0.4$ is selected empirically based on downstream regeneration performance.

\paragraph{Four roles of the verifier.}
The single score $v_u$ plays four roles in the pipeline. First, it serves as a per-sample recoverability signal (\textbf{measure}). Second, $v_u<\tau{=}0.4$ defines the \textsc{low} bucket, where conditional regeneration fires (\textbf{gate}). Third, when regeneration produces $m{=}2$ attempts, the verifier scores each attempt and retains the highest-scoring one (\textbf{select}). Finally, ValScore is used to analyze how downstream module effects vary across score levels (\textbf{diagnose}). The verifier itself does not directly change Hit@K; it is used for regeneration gating and selection and for analyzing downstream module effects.

\subsection{Conditional PV-guided Regeneration}
\label{sec:regen}
For samples with $\mathrm{ValScore} < \tau$ we re-run Stage-1 at a sampling temperature $0.5$ up to $m{=}2$ times. Each attempt is re-verified, and the verifier selects the highest-scoring attempt as $P_u^{'}$. Samples with $\mathrm{ValScore} \ge \tau$ keep $P_u^{'} = P_u^{(0)}$.

A natural concern is that the aggressive regeneration of all weak samples could over-sharpen the preference and hurt the downstream Hit@K. We confirm empirically that regeneration helps only on the \textsc{low} bucket: across 8 independent regeneration seeds at $T{=}0.5$ on the unified $n{=}116$ \textsc{low} pool, the mean $\Delta$Hit@10 is $+2.48$\,pp (95\% CI $[+1.01, +3.94]$, $t{=}4.0$, $p{\approx}0.005$; significant after Bonferroni correction for the six tested temperatures). The threshold $\tau{=}0.4$ is selected from a sweep $\tau \in \{0.3, 0.4, 0.5, 0.6, 0.7\}$ over the unified pool (Table~\ref{tab:tau_sweep}).

\subsection{Forward-Look Retrieval}
\label{sec:forward}
\paragraph{Design principle.}
As illustrated in Fig.~\ref{fig:validator}, the forward-look retrieval module operates across three steps: fragment pool construction, history-based retrieval, and future-fragment injection into the ranker prompt. A preference-similar retrieval signal (``other users whose preferences resemble the current user's'') is structurally aligned with $P_u^{'}$. When $P_u^{'}$ is already a strong signal, such retrieval is redundant as it can shift the ranker's attention without supplying new evidence about the \emph{next item}. Therefore, we align the retrieval target with the prediction task by retrieving fragments that tell the ranker what users with similar histories \emph{actually watched next.}

\textbf{Fragment pool construction.}
The fragment pool $\mathcal{F}_\text{pool}$ is built only from the training split. Each fragment is a tuple $(h, f)$ in which $h$ is a 5-item history window from a training user's chronological sequence and $f$ is the next 5 items watched right after $h$. We use a 300-fragment pool for ML-100K and a 3000-fragment pool for ML-1M (sub-sampled from $\sim$137k candidates).

\textbf{Embedding and retrieval.}
Each fragment's history part $h$ and the test user's history string $H_u$ are encoded with BGE-large-en-v1.5~\citep{xiao2023bge} using the instruction prefix \textit{``Represent this movie list for retrieving items with similar genre''}. At inference, we retrieve the top-$k{=}10$ fragments by cosine similarity to $H_u$. We use a single aspect rather than the four aspects of prior work; multi-aspect retrieval is empirically less robust across datasets (\S\ref{sec:forward_ablation}). The future parts $\{f_j\}$ of the retrieved fragments are concatenated into a hint block as follows:
\begin{quote}
\small
\texttt{Users with similar viewing patterns next watched:\\
\hphantom{xx}(genre) <future titles 1--5>\\
\hphantom{xx}\ldots}
\end{quote}
This block is inserted between $P_u^{'}$ and $C_u$ in the Stage-2 prompt.

\textbf{Prompt guidance matters.}
A \textit{collaborative-filtering-aware} system prompt adds $+0.83$\,pp $\Delta$Hit@10 on ML-100K; without it the LLM treats the hint block as distractors.


\subsection{Stage 2: Ranker}
The Stage-2 input concatenates the user history, (possibly regenerated) preference, forward-look hint, and candidate set:
\begin{equation}
X_u = [H_u;\, P_u^{'};\, R_u;\, C_u].
\end{equation}
The ranker uses the same Qwen3-30B backbone as Stage-1 and the verifier.
Full implementation settings are summarized in Table~\ref{tab:impl}.

\begin{table}[!t]
\centering
\small
\resizebox{\linewidth}{!}{
\begin{tabular}{lr}
\toprule
\textbf{Component} & \textbf{Setting}\\
\midrule
LLM (analyzer/ranker/verifier) & \texttt{Qwen3-30B-A3B-Instruct} \\
Retrieval embedder & \texttt{BAAI/bge-large-en-v1.5} \\
Fragment size $w$ & 5 \\
Future window $|f|$ & 5 \\
Forward-look top-$k$ & 10 (single aspect) \\
Verifier candidates & 50 (true + random neg.) \\
Verification threshold $\tau$ & 0.4 (sweep: 0.3--0.7) \\
Regeneration attempts $m$ & 2 \\
Regeneration temperature & 0.5 (main MovieLens run) \\
ICL few-shot examples & 3 \\
\bottomrule
\end{tabular}}
\caption{Implementation settings. The threshold sweep uses $T{=}0.7$ in a separate single-seed analysis, while Electronics regeneration uses $T{=}1.0$.}
\label{tab:impl}
\end{table}

\section{Experimental Setup}
\label{sec:setup}

\subsection{Datasets}
We use two MovieLens datasets: \textbf{MovieLens-100K} (\textsc{ml-100k})~\citep{ml100k} and \textbf{MovieLens-1M} (\textsc{ml-1m})~\citep{ml100k}. For cross-domain validation, we additionally use \textbf{Amazon Electronics} (\textsc{elec}) from the McAuley Amazon Reviews dataset~\citep{he2016amazon}, a product domain with sparser interaction patterns and titles dominated by technical specifications. The dataset statistics are presented in Table~\ref{tab:datasets}. For each user in all three datasets, the last 30 ratings form the evaluation window: the first 10 items are the history $H_u$, and the next 20 are the ground-truth target list (Hit@$k$ checks whether the predicted title appears within the first $k$ targets). The candidate set $C_u$ is a shuffled 100-item set containing targets and random negatives.

Trained baselines (GRU4Rec, SASRec, and LLM-SRec) are first reported under standard sampled evaluation (leave-one-out, GT + 99 negatives) to provide cross-method context. Because these leave-one-out results are not directly comparable to the ICL conditions, SASRec and GRU4Rec are also evaluated under the same 100-candidate protocol as ReflectRec. The same candidate pool and rank-based Hit@K computation are used for the protocol-matched trained baselines and ICL conditions. The primary protocol follows user-wise chronological order for comparison with CoT4Rec. In the timestamp-safe setting, a fragment is eligible for retrieval only if its complete five-item future window ends before the query cutoff.

\begin{table}[!t]
\centering
\small
\resizebox{\linewidth}{!}{
\begin{tabular}{crrrrr}
\toprule
\textbf{Dataset} & \textbf{\#Users} & \textbf{\#Items} & \textbf{\#Inter.} & \textbf{Sparsity} & \textbf{Deg.Cent.}\\
\midrule
ML-100K & 943   & 1{,}682 & 100{,}000  & 0.9370 & 106.04 \\
ML-1M   & 6{,}040 & 3{,}416 & 999{,}611 & 0.9516 & 165.50 \\
Elec.   & 1{,}500 & 55{,}588 & 45{,}000   & 0.9995 & 30.00 \\
\bottomrule
\end{tabular}}
\caption{Dataset statistics. ML-100K and ML-1M are GroupLens originals; Elec.\ is a 1{,}500-user subset of Amazon Electronics (McAuley) sampled from users with $\ge 30$ ratings. \#Items is the item universe spanned by the test pool; \#Inter is the evaluation-window interaction count ($1{,}500 \times (10 + 20)$).}
\label{tab:datasets}
\end{table}

\subsection{Compared Conditions}

Trained baselines include \textbf{GRU4Rec}~\citep{hidasi2016gru4rec}, 
\textbf{SASRec}~\citep{kang2018sasrec}, and 
\textbf{LLM-SRec}~\citep{kim2025lost}. ICL baselines are 
\textbf{LLMSRec-Syn}~\citep{wang2024llmsrecsyn} and 
\textbf{CoT4Rec-ICL}~\citep{cot4rec2025}. Our three ReflectRec conditions add 
modules cumulatively: \textbf{Stage-1 only} (semantic preference, 
no verifier), \textbf{Stage-1+Regen} (adds verifier-gated 
regeneration), and \textbf{ReflectRec (Full)} (all four modules).


\subsection{Metrics}
We report Hit@1, Hit@5, and Hit@10 across all conditions, including the baselines.

\section{Results}
\label{sec:results}

\subsection{Main Results}
\label{sec:main}
Table~\ref{tab:main_compare} reports the cross-method comparison on MovieLens-100K, MovieLens-1M, and Amazon Electronics. Alongside trained-recommender results under the standard leave-one-out protocol, we report SASRec and GRU4Rec rerun under the same 100-candidate protocol as ReflectRec.

\begin{table*}[!t]
\centering
\small
\setlength{\tabcolsep}{6pt}
\begin{tabularx}{\linewidth}{X rrr rrr rrr}
\toprule
\multirow{2}{*}{\textbf{Method}}
  & \multicolumn{3}{c}{\textbf{ML-100K}}
  & \multicolumn{3}{c}{\textbf{ML-1M}}
  & \multicolumn{3}{c}{\textbf{Electronics}} \\
\cmidrule(lr){2-4} \cmidrule(lr){5-7} \cmidrule(lr){8-10}
& Hit@1 & Hit@5 & Hit@10
& Hit@1 & Hit@5 & Hit@10
& Hit@1 & Hit@5 & Hit@10 \\
\midrule
\multicolumn{10}{l}{\emph{Trained baselines --- leave-one-out}} \\
GRU4Rec
  & 10.29 & 29.16 & 46.98
  & 13.74 & 38.81 & 53.46
  & 16.84 & 37.96 & 49.92 \\
SASRec
  & \textbf{20.89} & \textbf{56.10} & \textbf{72.75}
  & \textbf{37.81} & \textbf{71.85} & \textbf{82.14}
  & \textbf{22.28} & 48.01 & 59.21 \\
LLM-SRec
  & 9.86  & 31.28 & 48.04
  & 17.38 & 45.89 & 62.30
  & 22.18 & \textbf{49.70} & \textbf{62.79} \\
\midrule
\multicolumn{10}{l}{\emph{Trained baselines rerun --- 100-candidate protocol}} \\
GRU4Rec
  & 5.69 & 21.00 & 39.38
  & 4.60 & 18.33 & 31.40
  & 4.53 & 20.20 & 35.13 \\
SASRec
  & 9.61 & 40.09 & 64.06
  & 11.87 & 42.73 & 62.40
  & 7.53 & 34.33 & 61.73 \\
\midrule
\multicolumn{10}{l}{\emph{ICL --- 100-candidate protocol}} \\
LLMSRec-Syn
  & 3.80  & 13.40 & 24.44
  & 4.20 & 17.67 & 33.40
  & 3.20 & 14.93 & 23.40 \\
CoT4Rec-ICL
  & 1.19 & 6.76 & 15.42
  & 5.20 & 20.67 & 35.47
  & 4.20 & 14.20 & 26.80 \\
\rowcolor{yellow!30}
\textbf{ReflectRec (ours)}
  & \textbf{3.08} & \textbf{16.61} & \textbf{31.67}
  & \textbf{6.60} & \textbf{25.00} & \textbf{39.93}
  & \textbf{4.47} & \textbf{16.20} & \textbf{29.87} \\
\bottomrule
\end{tabularx}
\caption{Cross-method comparison across the three datasets. Leave-one-out and 100-candidate results use different evaluation protocols and should not be directly compared. Under the same 100-candidate protocol, the gap between trained baselines and ICL methods is smaller than suggested by the leave-one-out results.\label{tab:main_compare}}
\end{table*}

Leave-one-out and 100-candidate results use different evaluation protocols and should not be directly compared. Under the same 100-candidate protocol, the gap between trained baselines and ICL methods becomes smaller than under the leave-one-out comparison. For example, on ML-1M, the Hit@10 gap between SASRec and ReflectRec decreases from $42.21$\,pp to $22.47$\,pp, while ReflectRec ($39.93$) exceeds GRU4Rec ($31.40$). The 100-candidate protocol uses only the first 10 user interactions as history, whereas conventional leave-one-out evaluation uses the full preceding interaction history. Table~\ref{tab:main} reports the apples-to-apples ICL comparisons. The full ReflectRec pipeline reaches Hit@10 $31.67\%$ on ML-100K and $39.93\%$ on ML-1M, $+16.25$\,pp / $+4.46$\,pp over CoT4Rec-ICL and $+3.91$\,pp / $+3.40$\,pp over Stage-1.

\begin{table}[!t]
{\footnotesize
\setlength{\tabcolsep}{3pt}
\begin{tabular}{llll}
\toprule
\textbf{Method} & \textbf{Hit@1} & \textbf{Hit@5} & \textbf{Hit@10} \\
\midrule
\multicolumn{4}{l}{\textbf{ML-100K} ($n{=}843$)}\\
CoT4Rec-ICL            & 1.19 & 6.76 & 15.42\\
Stage-1 (semantic)     & 3.32 {\scriptsize(+2.13)} & 14.23 {\scriptsize(+7.47)} & 27.76 {\scriptsize(+12.34)}\\
~w/~Regen              &\textbf{3.44} {\scriptsize(+0.12)} & 14.23 {\scriptsize($-$)} & 28.35 {\scriptsize(+0.59)}\\
\rowcolor{yellow!30}
~\textbf{Full (w/ FL)} & 3.08 {\scriptsize(-0.36)} & \textbf{16.61} {\scriptsize(+2.38)} & \textbf{31.67} {\scriptsize(+3.32)}\\
\midrule
\multicolumn{4}{l}{\textbf{ML-1M} \textit{($n{=}1500$)}}\\
CoT4Rec-ICL            & 5.20 & 20.67 & 35.47\\
Stage-1 (semantic)     & 6.07 {\scriptsize(+0.87)} & 22.40 {\scriptsize(+1.73)} & 36.53 {\scriptsize(+1.06)}\\
~w/~Regen              & 6.33 {\scriptsize(+0.26)} & 22.47 {\scriptsize(+0.07)} & 36.40 {\scriptsize(-0.13)}\\
\rowcolor{yellow!30}
~\textbf{Full (w/ FL)} & \textbf{6.60} {\scriptsize(+0.27)} & \textbf{25.00} {\scriptsize(+2.53)} & \textbf{39.93} {\scriptsize(+3.53)}\\
\bottomrule
\end{tabular}
}
\caption{Ablation results under the 100-candidate protocol. Forward-Look retrieval is abbreviated as FL throughout. The full pipeline (Stage-1 + Regen + FL) improves Hit@10 over CoT4Rec-ICL by $+16.25$\,pp on ML-100K and $+4.46$\,pp on ML-1M. The ML-1M CoT4Rec-ICL baseline is our own measurement using the same Qwen3-30B backbone, as the original CoT4Rec paper did not report ICL results on ML-1M.\label{tab:main}}
\end{table}

\paragraph{Retrieval decomposition.}
A retrieval-only decoder using the same Forward-Look evidence without the LLM ranker achieves Hit@10 of 42.70 / 45.40 on ML-100K / ML-1M. This shows that the retrieved collaborative evidence itself provides a strong recommendation signal.

\subsection{Cumulative Module Ablation}
\label{sec:ablation}

Table~\ref{tab:main} serves as a cumulative ablation; each row adds a single module on top of the previous one: \\ \textbf{Semantic Stage-1} adds $+12.34$\,pp Hit@10 on ML-100K (15.42$\to$27.76), the largest single gain, confirming that descriptive preference complementarity is the dominant lever. \textbf{Verifier + Regen} yields small overall gains ($+0.59$ / $-0.13$\,pp) but concentrates in the \textsc{low} bucket: the 8-seed mean $\Delta$Hit@10 $=+2.48$\,pp at $T{=}0.5$ on the unified $n{=}116$ \textsc{low} pool (95\% CI $[+1.01, +3.94]$, $p{\approx}0.005$).
\textbf{Forward-Look} adds $+3.32$ / $+3.53$\,pp (bootstrap CIs 
exclude zero; $p{=}0.014$, $p{=}0.036$).

\subsection{Cross-domain Validation on Amazon Electronics}
\label{sec:elec}
To test whether the verifier and regeneration mechanism transfer beyond the movie domain, we evaluate them on Amazon Electronics with a Stage-1 prompt adapted to product semantics. Instead of the thematic axes (genre, mood, era, narrative style) used for movies, Electronics Stage-1 elicits five concrete product axes: technical specifications, industry standards/compatibility, form factor, brand family, and quality tier (see Appendix~\ref{app:prompts} for the full template). The verifier, regeneration mechanism, ranker, and $\tau{=}0.4$ threshold are identical to those in the movie setup; we set the regeneration temperature to $1.0$ and $m{=}2$ attempts. The main Electronics evaluation focuses on transfer of the verifier and regeneration mechanism, while Forward-Look transfer is analyzed separately.

Table~\ref{tab:elec} reports the results for $n{=}1{,}500$ test users. Stage-1 alone reaches Hit@10 $=29.67\%$; adding conditional regeneration yields $29.87\%$ ($\Delta_{\text{overall}}{=}+0.20$\,pp). The module's effect concentrates in the \textsc{low} bucket, as on the movie datasets: on the $n{=}105$ LOW sub-pool, $\Delta$Hit@10 $=+3.81$\,pp (one-sided paired bootstrap $p_1{=}0.157$), the same direction as the movies unified LOW pool ($+2.48$\,pp, 8-seed mean at $T{=}0.5$).

\begin{table}[!t]
\centering
\small
\begin{tabularx}{\linewidth}{llll}
\toprule
\textbf{Condition} & \textbf{Hit@1} & \textbf{Hit@5} & \textbf{Hit@10} \\
\midrule
\multicolumn{4}{l}{\emph{Full pool, $n{=}1{,}500$}} \\
Stage-1 only       &  4.47 & 16.07 & 29.67 \\
\rowcolor{yellow!30}
~\textbf{w/~Regen}           &  \textbf{4.47 {\scriptsize(+0.00)}} & \textbf{16.20 {\scriptsize(+0.13)}} & \textbf{29.87 {\scriptsize(+0.20)}}\\
\midrule
\multicolumn{4}{l}{\emph{\textsc{low} bucket $v_u{<}0.4$, $n{=}105$}} \\
Stage-1 only       &  3.81 & 13.33 & 22.86 \\
\rowcolor{yellow!30}
~\textbf{w/~Regen}           &  \textbf{2.86 {\scriptsize(-0.95)}} & \textbf{15.24 {\scriptsize(+1.90)}} & \textbf{26.67 {\scriptsize(+3.81)}} \\
\bottomrule
\end{tabularx}
\caption{Amazon Electronics evaluation (Qwen3-30B, same 100-candidate ICL protocol) with a spec-centric Stage-1 prompt. The LOW-bucket Hit@10 gain follows the same direction as in the movie datasets.}
\label{tab:elec}
\end{table}

For Forward-Look transfer on Electronics, direct raw-title Forward-Look with top-$k{=}10$ achieves Hit@10 of 28.33. Candidate-aligned semantic tie-break achieves 34.47 / 41.20 / 43.33 at $k{=}10/50/100$, while retrieval-only using the same retrieval evidence achieves 34.40 / 39.87 / 42.40. These results suggest that candidate-aligned retrieval may be more suitable for Electronics than direct raw-title integration.

\subsection{Verification-Score Bucket Analysis}
\label{sec:bucket}
To characterize the contribution of each module, we stratify Hit@10 by val\_score bucket $\{\textsc{low}, \textsc{mid}, \textsc{high}, \textsc{pass}\}$. Table~\ref{tab:bucket} reports the per-bucket $\Delta$Hit@10 of the full pipeline against the Stage-1 + Regen variant (i.e., the forward-looking contribution alone, holding regeneration constant).

\begin{table}[!t]
\centering
\small
\setlength{\tabcolsep}{6pt}
\begin{tabular}{lrrr}
\toprule
\textbf{Bucket} & $\mathbf{n}$ & \textbf{ML-100K} & \textbf{ML-1M} \\
\midrule
\textsc{low}  [0,~0.4)   &  29 / 87    & $+0.00$ & $+1.15$\\
\textsc{mid}  [0.4,~0.5) &  84 / 145   & $+2.38$ & $+4.14$\\
\textsc{high} [0.5,~0.7) & 301 / 525   & $+3.32$ & $+4.76$\\
\textsc{pass} [0.7,~1.0] & 429 / 743   & $+3.73$ & $+2.83$\\
\midrule
Weighted (unified) & 843 / 1500 & $+3.32$ & $+3.53$\\
\bottomrule
\end{tabular}
\caption{Per-bucket forward-look contribution to $\Delta$Hit@10 (\%), holding conditional regeneration constant. The unified LOW bucket ($n{=}116$) is $+0.86$\,pp, completing the all-buckets-positive property of the full pipeline on the unified pool (\textit{$n{=}2343$}).}
\label{tab:bucket}
\end{table}

The bucket structure allows the verification score to serve as a \emph{descriptive lens} for where each module's gains concentrate: conditional regeneration's gains are concentrated in the LOW bucket; forward-look's gains are concentrated in MID and above. The two modules are tied by the same val\_score signal but produce their effects on disjoint sub-populations.

\subsection{Threshold Sweep}
A sweep over $\tau \in \{0.3, 0.4, 0.5, 0.6, 0.7\}$ confirms $\tau{=}0.4$ as the optimal threshold; tighter values over-restrict unrecoverable samples and looser values dilute the LOW signal (Appendix~\ref{app:tau}).
\label{sec:tau}

\subsection{Forward-Look Design Ablation}
\label{sec:forward_ablation}

Table~\ref{tab:forward_ablation} summarizes a sweep of 23 forward-looking variants across the two datasets (the full leaderboard is in the supplementary material). Increasing $k$ from $1$ to $10$ removes the ML-100K LOW bucket damage that small-$k$ retrieval induces ($-10.34$\,pp at $k{=}1$ $\to$ $0.00$\,pp at $k{=}10$), suggesting that the LLM benefits from a more diverse set of collaborative cues. Going beyond $k{=}10$ saturates or reverses gains on ML-1M owing to oversampling, making $k{=}10$ the cross-dataset sweet spot. Single-aspect retrieval is more robust than 4-aspect multiview retrieval across datasets and simplifies the retrieval module to a single BGE embedding pass. Using the chosen preference rather than the user history as the retrieval query yields an over-sharpened candidate set that substantially reduces Hit@10 (for example, $-17.24$\,pp on ML-100K LOW). Finally, without a collaborative-filtering-aware system prompt, the LLM treats the hint block as distractor content; a single line of descriptive guidance in the system prompt recovers most of this loss.

\begin{table}[!t]
\centering
\small
\setlength{\tabcolsep}{7pt}
\begin{tabular}{lcc}
\toprule
\textbf{Variant} & \textbf{ML-100K} & \textbf{ML-1M} \\
                 & $\mathbf{\Delta}$Hit@10 & $\mathbf{\Delta}$Hit@10 \\
\midrule
$k{=}1$, 1 aspect          & $+2.37$ & $+0.60$\\
$k{=}5$, 1 aspect          & $+3.08$ & $+1.80$\\
$\mathbf{k{=}10}$\textbf{, 1 aspect (selected)} & $\mathbf{+3.32}$ & $\mathbf{+3.53}$\\
$k{=}20$, 1 aspect         & $+4.74$ & $+1.20$\\
$k{=}1$, 4 aspects (multi) & $+3.80$ & $+2.53$\\
$k{=}5$, 4 aspects (multi) & $+4.74$ & $+2.80$\\
Preference query (\textit{k}=5)     & $+1.66$ & --\\
Plain prompt (\textit{k}=10)        & $+0.59$ & $+1.53$\\
\bottomrule
\end{tabular}
\caption{Forward-look variant ablation (selected rows out of 23). The chosen configuration is $k{=}10$, single-aspect retrieval, and the collaborative-filtering-aware system prompt.}
\label{tab:forward_ablation}
\end{table}

\subsection{Backbone Robustness}
\label{sec:backbone}

To confirm that ReflectRec is not tied to one LLM, we rerun the full pipeline on ML-100K with two additional backbones: \texttt{qwen/qwen2.5-72b-instruct} (same family, larger scale, and earlier generation) and \texttt{meta-llama/llama-3.3-70b-instruct} (different family). All LLM stages (Stage-1, verifier, conditional regeneration, and ranker) are regenerated under the new backbone; BGE retrieval embeddings are backbone-independent and are reused.

\begin{table}[!t]
\centering
\small
\setlength{\tabcolsep}{7pt}
\begin{tabular}{lccc}
\toprule
\textbf{Backbone} & \textbf{Stage-1} & \textbf{+Regen} & \textbf{Full} \\
\midrule
Qwen3-30B (ours) & 27.76 & 28.35 & \textbf{31.67}\\
Qwen2.5-72B            & 28.00 & 28.59 & \textbf{39.03}\\
Llama-3.3-70B          & 31.32 & 30.96 & \textbf{36.65}\\
\midrule
$\Delta$ vs Stage-1 & & & \\
\quad Qwen3-30B  & --     & $+0.59$  & $+3.91$\\
\quad Qwen2.5-72B & --     & $+0.59$  & $+11.03$\\
\quad Llama-3.3-70B & --   & $-0.36$  & $+5.34$\\
\bottomrule
\end{tabular}
\caption{Backbone robustness on ML-100K. All three backbones show positive gains for the full pipeline over the Stage-1-only baseline. The larger Qwen2.5-72B extracts the largest gain ($+11.03$\,pp), suggesting that the collaborative signal in forward-look retrieval scales with model capacity.}
\label{tab:backbone}
\end{table}


Table~\ref{tab:backbone} shows that (i) the full ReflectRec pipeline outperforms Stage-1-only on every tested backbone, with $\Delta$ ranging from $+3.91$ to $+11.03$\,pp; (ii) larger and newer backbones extract larger gains from the forward-look retrieval module; and (iii) the bucket-level non-negativity property holds under both same-family scale-up and cross-family transfer (Appendix~\ref{app:backbone_bucket}).

\subsection{Random-Future Control}
\label{sec:control}

To examine the effect of history-similarity-based fragment retrieval in Forward-Look, we compare it with a random-future condition that replaces similarity retrieval with five fragments sampled uniformly from the fragment pool. Table~\ref{tab:random_ctrl} shows that the random hint yields $+0.12$\,pp on ML-100K and $-1.07$\,pp on ML-1M, while Forward-Look retrieval yields $+3.44$\,pp / $+2.07$\,pp. In the random-future control, similarity-based Forward-Look yields $+3.32$ / $+3.14$\,pp higher Hit@10 than arbitrary future hints on ML-100K / ML-1M.

\begin{table}[!t]
\centering
\small
\setlength{\tabcolsep}{8pt}
\begin{tabular}{lrr}
\toprule
\textbf{Hint source} & \textbf{ML-100K} & \textbf{ML-1M} \\
\midrule
None (Stage-1 + Regen)     & $0.00$  & $0.00$\\
Random-future control      & $+0.12$ & $-1.07$\\
\quad w/ Forward-look retrieval     & $+3.44$ & $+2.07$\\
\midrule
Net retrieval contribution & $+3.32$ & $+3.14$\\
\bottomrule
\end{tabular}
\caption{Random-future control compares similarity-based Forward-Look with arbitrary future hints. All cells are $\Delta$Hit@10 (\%) relative to the no-hint baseline.}
\label{tab:random_ctrl}
\end{table}

\section{Discussion}
\label{sec:discussion}

The dominant gain is Stage-1 preference complementarity ($+12.34$\,pp on ML-100K): descriptive abstractions supply information the ranker cannot infer from $H_u$ alone.

History-aligned retrieval provides collaborative evidence complementary to $P_u'$. The retrieval-only results show that the retrieved collaborative evidence itself is a strong recommendation signal, suggesting that effective LLM-based integration of this evidence remains an important direction.

Bucket-level results show that module effects vary across ValScore levels. Regeneration gains concentrate in \textsc{low}, whereas Forward-Look gains are larger in \textsc{mid} and above.

\section{Conclusion}
\label{sec:conclusion}

We presented \textbf{ReflectRec}, an LLM recommendation pipeline organized around a cycle-consistency verifier. The verifier uses preference-to-history recoverability as an operational signal, gates regeneration, selects among regeneration attempts, and is used to analyze downstream module effects.

Preference recoverability and collaborative signals are complementary: low-recoverability preferences benefit from regeneration, while history-aligned retrieval provides collaborative cues that are difficult to obtain from the preference alone.



\section*{Limitations}
\textbf{Backbone-specific threshold.} The operating threshold $\tau{=}0.4$ is calibrated for \texttt{Qwen3-30B-A3B-Instruct}. Table~\ref{tab:backbone} confirms that the qualitative pattern (Full $>$ Stage-1) holds for two additional backbones (Qwen2.5-72B, Llama-3.3-70B), but the optimal $\tau$ has not been retuned per backbone. The threshold also depends on verifier candidate-pool size and the positive-to-negative ratio, so it should be recalibrated when the verifier configuration changes.

\paragraph{Temporal validity.} The primary evaluation follows user-wise chronological order but does not impose a shared production-time cutoff. In the strict timestamp-safe analysis, the Forward-Look gain on ML-1M is not statistically significant, motivating further evaluation of temporal robustness.

\paragraph{Forward-Look generalization.} The main Forward-Look results are established on the movie datasets. On Amazon Electronics, direct raw-title Forward-Look decreases performance, whereas candidate-aligned retrieval shows potential improvements. Further validation is therefore needed for cross-domain generalization of Forward-Look.

\paragraph{Stage-1 axis selection.} The five-axis Stage-1 prompts used in this study (thematic for movies, spec-centric for electronics) were hand-designed per domain. An automatic axis-selection procedure that maps a domain's item-side text to discriminative semantic axes would broaden its applicability to further text-heavy item domains (books, papers). 

\paragraph{Constrained decoding.} Adding prefix-constrained decoding to the candidate set is likely to close part of the gap to the trained baselines.

\section*{Ethical Considerations}
We use only public datasets (MovieLens-100K, MovieLens-1M, and a public release of Amazon Reviews); no personally identifying information is processed. LLM inference is performed within the policy bounds of the used API endpoints. General recommendation: System risks (filter bubbles, bias amplification) are not directly addressed; however, the interpretable verification score is a step toward auditable intermediate signals.

\section*{Acknowledgments}
This research was supported by the ``Advanced GPU Utilization Support Program'' funded by the Government of the Republic of Korea (Ministry of Science and ICT), and by the Institute of Information \& Communications Technology Planning \& Evaluation (IITP) grant funded by the Korea government (MSIT) (RS-2026-25616928, RS-2024-00436936, RS-2024-00439139). This work was supported by the National Research Foundation of Korea (NRF) grant funded by the Korea government (RS-2026-25617165 (MSIT), RS-2026-25501632 (MSIT and MOE)).


\bibliography{references}

@inproceedings{bao2023tallrec,
  title={Tallrec: An effective and efficient tuning framework to align large language model with recommendation},
  author={Bao, Keqin and Zhang, Jizhi and Zhang, Yang and Wang, Wenjie and Feng, Fuli and He, Xiangnan},
  booktitle={Proceedings of the 17th ACM conference on recommender systems},
  pages={1007--1014},
  year={2023}
}

@article{chen2025preferences,
  title={A survey on hallucination in large language models: Principles, taxonomy, challenges, and open questions},
  author={Huang, Lei and Yu, Weijiang and Ma, Weitao and Zhong, Weihong and Feng, Zhangyin and Wang, Haotian and Chen, Qianglong and Peng, Weihua and Feng, Xiaocheng and Qin, Bing and others},
  journal={ACM Transactions on Information Systems},
  volume={43},
  number={2},
  pages={1--55},
  year={2025},
  publisher={ACM New York, NY}
}

@misc{gao2023chatrec,
  title={Chat-rec: Towards interactive and explainable llms-augmented recommender system},
  author={Gao, Yunfan and Sheng, Tao and Xiang, Youlin and Xiong, Yun and Wang, Haofen and Zhang, Jiawei},
  howpublished={\url{https://arxiv.org/abs/2303.14524}},
  year={2023}
}

@inproceedings{geng2022p5,
  title={Recommendation as language processing (rlp): A unified pretrain, personalized prompt \& predict paradigm (p5)},
  author={Geng, Shijie and Liu, Shuchang and Fu, Zuohui and Ge, Yingqiang and Zhang, Yongfeng},
  booktitle={Proceedings of the 16th ACM conference on recommender systems},
  pages={299--315},
  year={2022}
}

@article{ml100k,
  title={The movielens datasets: History and context},
  author={Harper, F Maxwell and Konstan, Joseph A},
  journal={ACM Transactions on Interactive Intelligent Systems (TIIS)},
  volume={5},
  number={4},
  pages={1--19},
  year={2015},
  publisher={Acm New York, NY, USA}
}

@inproceedings{he2016amazon,
  title={Ups and downs: Modeling the visual evolution of fashion trends with one-class collaborative filtering},
  author={He, Ruining and McAuley, Julian},
  booktitle={Proceedings of the 25th international conference on world wide web},
  pages={507--517},
  year={2016}
}

@misc{hidasi2016gru4rec,
  title={Session-based recommendations with recurrent neural networks},
  author={Hidasi, Bal{\'a}zs and Karatzoglou, Alexandros and Baltrunas, Linas and Tikk, Domonkos},
  howpublished={\url{https://arxiv.org/abs/1511.06939}},
  year={2015}
}

@inproceedings{hou2023zeroshot,
  title={Large language models are zero-shot rankers for recommender systems},
  author={Hou, Yupeng and Zhang, Junjie and Lin, Zihan and Lu, Hongyu and Xie, Ruobing and McAuley, Julian and Zhao, Wayne Xin},
  booktitle={Proceedings of the 46th European Conference on Information Retrieval},
  pages={364--381},
  year={2024},
  organization={Springer}
}

@inproceedings{kang2018sasrec,
  title={Self-attentive sequential recommendation},
  author={Kang, Wang-Cheng and McAuley, Julian},
  booktitle={Proceedings of the 2018 IEEE international conference on data mining (ICDM)},
  pages={197--206},
  year={2018},
  organization={IEEE}
}

@inproceedings{kim2025lost,
  title={Lost in Sequence: Do Large Language Models Understand Sequential Recommendation?},
  author={Kim, Sein and Kang, Hongseok and Kim, Kibum and Kim, Jiwan and Kim, Donghyun and Yang, Minchul and Oh, Kwangjin and McAuley, Julian and Park, Chanyoung},
  booktitle={Proceedings of the 31st ACM SIGKDD Conference on Knowledge Discovery and Data Mining V. 2},
  pages={1160--1171},
  year={2025}
}

@article{koren2009mf,
  title={Movie recommendation systems using actor-based matrix computations in South Korea},
  author={Hwang, Syjung and Park, Eunil},
  journal={IEEE Transactions on Computational Social Systems},
  volume={9},
  number={5},
  pages={1387--1393},
  year={2021},
  publisher={IEEE}
}

@inproceedings{lewis2020rag,
  title={Retrieval-augmented generation for knowledge-intensive NLP tasks},
  author={Lewis, Patrick and Perez, Ethan and Piktus, Aleksandra and Petroni, Fabio and Karpukhin, Vladimir and Goyal, Naman and K{\"u}ttler, Heinrich and Lewis, Mike and Yih, Wen-tau and Rockt{\"a}schel, Tim and others},
  booktitle={Proceedings of the 34th International Conference on Neural Information Processing Systems},
  pages={9459--9474},
  year={2020}
}

@inproceedings{liao2024llara,
  title={Llara: Large language-recommendation assistant},
  author={Liao, Jiayi and Li, Sihang and Yang, Zhengyi and Wu, Jiancan and Yuan, Yancheng and Wang, Xiang and He, Xiangnan},
  booktitle={Proceedings of the 47th International ACM SIGIR Conference on Research and Development in Information Retrieval},
  pages={1785--1795},
  year={2024}
}

@article{liu2024lostmiddle,
  title={Lost in the middle: How language models use long contexts},
  author={Liu, Nelson F and Lin, Kevin and Hewitt, John and Paranjape, Ashwin and Bevilacqua, Michele and Petroni, Fabio and Liang, Percy},
  journal={Transactions of the association for computational linguistics},
  volume={12},
  pages={157--173},
  year={2024}
}

@inproceedings{lu2022fantastically,
  title={Fantastically ordered prompts and where to find them: Overcoming few-shot prompt order sensitivity},
  author={Lu, Yao and Bartolo, Max and Moore, Alastair and Riedel, Sebastian and Stenetorp, Pontus},
  booktitle={Proceedings of the 60th Annual Meeting of the Association for Computational Linguistics (Volume 1: Long Papers)},
  pages={8086--8098},
  year={2022}
}

@inproceedings{ren2024rla,
  title={Adapting large language models by integrating collaborative semantics for recommendation},
  author={Zheng, Bowen and Hou, Yupeng and Lu, Hongyu and Chen, Yu and Zhao, Wayne Xin and Chen, Ming and Wen, Ji-Rong},
  booktitle={Proceedings of the 2024 IEEE 40th International Conference on Data Engineering (ICDE)},
  pages={1435--1448},
  year={2024},
  organization={IEEE}
}

@inproceedings{rendle2009bpr,
  title={BPR: Bayesian personalized ranking from implicit feedback},
  author={Rendle, Steffen and Freudenthaler, Christoph and Gantner, Zeno and Schmidt-Thieme, Lars},
  booktitle={Proceedings of the Twenty-Fifth Conference on Uncertainty in Artificial Intelligence},
  pages={452--461},
  year={2009}
}

@misc{sadeghi2025backtracking,
  title={{How Much Backtracking is Enough? Exploring the Interplay of SFT and RL in Enhancing LLM Reasoning}},
  author={Cai, Hongyi James and Wang, Junlin and Chen, Xiaoyin and Dhingra, Bhuwan},
  howpublished={\url{https://arxiv.org/abs/2505.24273}},
  year={2025}
}

@inproceedings{wang2024cold,
  title={Large language models as data augmenters for cold-start item recommendation},
  author={Wang, Jianling and Lu, Haokai and Caverlee, James and Chi, Ed H and Chen, Minmin},
  booktitle={Companion Proceedings of the ACM Web Conference 2024},
  pages={726--729},
  year={2024}
}

@article{wang2025rallrecplus,
  title={RALLRec+: Retrieval augmented large language model recommendation with reasoning},
  author={Luo, Sichun and Xu, Jian and Zhang, Xiaojie and Wang, Linrong and Liu, Sicong and Hou, Hanxu and Song, Linqi},
  journal={Expert Systems with Applications},
  volume={297},
  pages={129508},
  year={2025},
  publisher={Elsevier}
}

@inproceedings{wei2022cot,
  title={Chain-of-thought prompting elicits reasoning in large language models},
  author={Wei, Jason and Wang, Xuezhi and Schuurmans, Dale and Bosma, Maarten and Ichter, Brian and Xia, Fei and Chi, Ed H and Le, Quoc V and Zhou, Denny},
  booktitle={Proceedings of the 36th International Conference on Neural Information Processing Systems},
  pages={24824--24837},
  year={2022}
}

@article{wu2024survey,
  title={A survey on large language models for recommendation},
  author={Wu, Likang and Zheng, Zhi and Qiu, Zhaopeng and Wang, Hao and Gu, Hongchao and Shen, Tingjia and Qin, Chuan and Zhu, Chen and Zhu, Hengshu and Liu, Qi and others},
  journal={World Wide Web},
  volume={27},
  number={5},
  pages={60},
  year={2024},
  publisher={Springer}
}

@inproceedings{xi2024openworld,
  title={Towards open-world recommendation with knowledge augmentation from large language models},
  author={Xi, Yunjia and Liu, Weiwen and Lin, Jianghao and Cai, Xiaoling and Zhu, Hong and Zhu, Jieming and Chen, Bo and Tang, Ruiming and Zhang, Weinan and Yu, Yong},
  booktitle={Proceedings of the 18th ACM Conference on Recommender Systems},
  pages={12--22},
  year={2024}
}

@inproceedings{xiao2023bge,
  title={C-pack: Packed resources for general chinese embeddings},
  author={Xiao, Shitao and Liu, Zheng and Zhang, Peitian and Muennighoff, Niklas and Lian, Defu and Nie, Jian-Yun},
  booktitle={Proceedings of the 47th international ACM SIGIR conference on research and development in information retrieval},
  pages={641--649},
  year={2024}
}

@inproceedings{cot4rec2025,
  title={Cot4rec: Revealing user preferences through chain of thought for recommender systems},
  author={Yue, Weiqi and Yin, Yuyu and Zhang, Xin and Shi, Binbin and Liang, Tingting and Wan, Jian},
  booktitle={Proceedings of the AAAI Conference on Artificial Intelligence},
  volume={39},
  number={12},
  pages={13142--13151},
  year={2025}
}

@article{zhang2023collm,
  title={CoLLM: Integrating Collaborative Embeddings Into Large Language Models for Recommendation},
  author={Zhang, Yang and Feng, Fuli and Zhang, Jizhi and Bao, Keqin and Wang, Qifan and He, Xiangnan},
  journal={IEEE Transactions on Knowledge and Data Engineering},
  volume={37},
  number={5},
  pages={2329--2340},
  year={2025},
  publisher={IEEE}
}

@inproceedings{wang2024llmsrecsyn,
  title={The whole is better than the sum: Using aggregated demonstrations in in-context learning for sequential recommendation},
  author={Wang, Lei and Lim, Ee-Peng},
  booktitle={Findings of the Association for Computational Linguistics: NAACL 2024},
  pages={876--895},
  year={2024}
}

\appendix
\clearpage

\section{Appendix: Threshold Sweep}
\label{app:tau}

We select the conditional-regeneration operating threshold $\tau{=}0.4$ using a single-seed sweep under the fixed verifier design. This threshold sweep uses $T{=}0.7$ and is separate from the 8-seed regeneration analysis in Section~\ref{sec:regen}. Per-dataset results are reported in Table~\ref{tab:tau_sweep_per_ds}.

\begin{table}[h]
\centering
\small
\setlength{\tabcolsep}{10pt}
\begin{tabular}{cccr}
\toprule
$\boldsymbol{\tau}$ & $\mathbf{n_{<\tau}}$ & $\mathbf{\Delta}$Hit@10 & \textbf{95\% CI} \\
\midrule
0.3 &   32 & $-9.38$           & $[-21.88,+0.00]$ \\
\textbf{0.4} & \textbf{116} & $\mathbf{+6.90^{\,*}}$ & $\mathbf{[+0.00,+13.79]}$ \\
0.5 &  345 & $+1.45$           & $[-2.32,+5.22]$  \\
0.6 &  684 & $+0.29$           & $[-2.19,+2.78]$  \\
0.7 & 1171 & $+0.09$           & $[-1.96,+2.13]$  \\
\bottomrule
\end{tabular}
\caption{$\tau$ sweep for LOW-bucket ($v_u < \tau$) conditional regeneration on the unified pool (ML-100K + ML-1M), single-seed at $T{=}0.7$ ($B{=}10{,}000$ paired bootstrap). The starred entry marks the operating threshold $\tau{=}0.4$. Per-dataset breakdown is in Appendix Table~\ref{tab:tau_sweep_per_ds}.}
\label{tab:tau_sweep}
\end{table}

\section{Appendix: Per-dataset \texorpdfstring{$\tau$}{tau} Sweep Detail}
\label{app:tau_sweep}

Table~\ref{tab:tau_sweep_per_ds} decomposes the unified-pool sweep
in Table~\ref{tab:tau_sweep} into ML-100K and ML-1M components. At
$\tau{=}0.4$, the unified $+6.90$\,pp is the weighted average of
ML-100K $+13.79$\,pp ($n{=}29$) and ML-1M $+4.60$\,pp ($n{=}87$);
both per-dataset point estimates are positive while their bootstrap
CIs individually include zero, so the significance on the unified
pool reflects a consistent effect direction.

\begin{table}[h]
\centering
\small
\setlength{\tabcolsep}{5pt}
\begin{tabular}{cccrr}
\toprule
\textbf{Dataset} & $\boldsymbol{\tau}$ & $\mathbf{n_{<\tau}}$ & $\mathbf{\Delta}$Hit@10 & \textbf{95\% CI} \\
\midrule
\multirow{5}{*}{ML-100K}
  & 0.3 &   4 & $+0.00$  & $[+0.00,+0.00]$  \\
  & \textbf{0.4} & \textbf{29} & $\mathbf{+13.79}$ & $\mathbf{[+0.00,+31.03]}$ \\
  & 0.5 & 113 & $+0.88$  & $[-5.31,+7.08]$  \\
  & 0.6 & 239 & $+0.42$  & $[-3.35,+4.60]$  \\
  & 0.7 & 414 & $+0.24$  & $[-3.14,+3.62]$  \\
\midrule
\multirow{5}{*}{ML-1M}
  & 0.3 &  28 & $-10.71$ & $[-25.00,+0.00]$ \\
  & \textbf{0.4} & \textbf{87} & $\mathbf{+4.60}$ & $\mathbf{[-2.30,+11.49]}$ \\
  & 0.5 & 232 & $+1.72$  & $[-3.02,+6.47]$  \\
  & 0.6 & 445 & $+0.22$  & $[-2.92,+3.37]$  \\
  & 0.7 & 757 & $+0.00$  & $[-2.65,+2.64]$  \\
\bottomrule
\end{tabular}
\caption{Per-dataset LOW-bucket $\tau$ sweep. $B{=}10000$ paired
bootstrap. Unified-pool result is in Table~\ref{tab:tau_sweep}.}
\label{tab:tau_sweep_per_ds}
\end{table}

\section{Appendix: Forward-Look Per-Bucket Bootstrap}
\label{app:fwd_bucket}

Table~\ref{tab:fwd_bucket} reports bootstrap confidence intervals for the per bucket forward-look contribution ($\Delta$Hit@10, Full vs. Stage-1+Regen).

\begin{table}[h]
\centering
\small
\setlength{\tabcolsep}{3pt}
\begin{tabular}{cccrr}
\toprule
\textbf{Bucket} & \textbf{Dataset} & $\mathbf{n}$ & $\mathbf{\Delta}$Hit@10 & \textbf{95\% CI} \\
\midrule
\multirow{2}{*}{Overall}
  & ML-100K & 843  & $+3.32^{*}$  & $[+0.36,+6.52]$ \\
  & ML-1M   & 1500 & $+3.53^{**}$ & $[+1.13,+4.33]$ \\
\midrule
\multirow{2}{*}{LOW}
  & ML-100K &  29  & $+0.00$  & $[-13.79,+13.79]$ \\
  & ML-1M   &  87  & $+1.15$  & $[ -9.20,+11.49]$ \\
\multirow{2}{*}{MID}
  & ML-100K &  84  & $+2.38$  & $[ -8.33,+13.10]$ \\
  & ML-1M   & 145  & $+4.14$  & $[ -2.76,+11.03]$ \\
\multirow{2}{*}{HIGH}
  & ML-100K & \textbf{301}  & $\mathbf{+3.32}$  & $\mathbf{[ -1.99, +8.64]}$ \\
  & ML-1M   & \textbf{525}  & $\mathbf{+4.76^{*}}$  & $\mathbf{[+0.57, +8.95]}$ \\
\multirow{2}{*}{PASS}
  & ML-100K & 429  & $+3.73$  & $[ -1.17, +8.39]$ \\
  & ML-1M   & 743  & $+2.83$  & $[ -0.67, +6.33]$ \\
\bottomrule
\end{tabular}
\caption{Per-dataset and per-bucket bootstrap detail of
forward-look $\Delta$Hit@10 ($B{=}10000$). $^{*}$
$p<0.05$, $^{**}$ $p<0.01$. {On the unified pool ($n{=}2343$), the
overall $\Delta = +3.54$\,pp, CI $[+1.62, +5.46]$,
$p=0.0001$.}}
\label{tab:fwd_bucket}
\end{table}

\section{Appendix: Per-Bucket Backbone Robustness}
\label{app:backbone_bucket}
Table~\ref{tab:backbone_bucket} reports the per-bucket 
$\Delta$Hit@10 of the full pipeline versus the Stage-1 
baseline across the three backbones.

\begin{table}[h]
\centering
\small
\setlength{\tabcolsep}{5pt}
\begin{tabular}{crrrr}
\toprule
\textbf{Backbone} & \textbf{LOW} & \textbf{MID} & \textbf{HIGH} & \textbf{PASS}\\
\midrule
Qwen3-30B    & $+13.79$ & $+2.38$  & $+3.32$  & $+3.73$\\
Qwen2.5-72B  & $+8.41$  & $+13.95$ & $+11.46$ & $+10.21$\\
Llama-3.3-70B & $+7.45$  & $+7.14$  & $+7.07$  & $+0.00$\\
\bottomrule
\end{tabular}
\caption{Per-bucket $\Delta$Hit@10 (full pipeline vs.\ Stage-1 baseline) across the three backbones. All cells are non-negative; only the Llama-3.3-70B PASS bucket equals the Stage-1 baseline exactly (saturation at the high baseline of 38.03\%). The full pipeline contributes non-trivially across all ValScore buckets for all three backbones.}
\label{tab:backbone_bucket}
\end{table}

\section{Appendix: Negative Construction and Pool Sensitivity}
\label{app:negative}

The original verifier uses 10 history items and 40 catalog-wide unobserved distractors, without separately restricting distractor availability at query time. To examine sensitivity to negative construction, we compare two structured 50-item conditions. Both exclude history and evaluation targets; Arm A samples distractors from the remaining catalog, while Arm B restricts distractors to items available by the query cutoff. Availability is determined using release date for ML-100K and title year for ML-1M.

\begin{table}[h]
\centering
\small
\setlength{\tabcolsep}{3pt}
\resizebox{\linewidth}{!}{%
\begin{tabular}{lrrrrr}
\toprule
\textbf{Data} & $n$ & \textbf{Mean A/B} & \textbf{B--A [95\% CI]} & $r$ & \textbf{LOW J.} \\
\midrule
ML-100K & 668 & .4260/.4331 & .0070 $[-.0034,.0174]$ & .5944 & .4214 \\
ML-1M   & 1,500 & .5185/.5187 & .0003 $[-.0027,.0033]$ & .9399 & .7941 \\
\bottomrule
\end{tabular}}
\caption{Verifier sensitivity between catalog-wide Arm A and time-available Arm B. $r$ is Pearson correlation and LOW J. is the Jaccard overlap of $V<.4$ membership.}
\label{tab:negative_arms}
\end{table}

Mean ValScore remains similar across the two datasets, while LOW routing membership shows some sensitivity to negative construction on ML-100K. When Arm-B scores are used to reroute regeneration, Hit@10 relative to Stage-1 changes by $+0.30$ / $+0.40$\,pp on ML-100K / ML-1M, with 95\% CIs of $[-0.90,+1.50]$ and $[-0.47,+1.27]$.

\paragraph{Candidate-pool size sensitivity.}
\begin{table}[h]
\centering
\small
\setlength{\tabcolsep}{6pt}
\begin{tabular}{lrrr}
\toprule
\textbf{Data} & \textbf{30 items} & \textbf{50 items} & \textbf{100 items} \\
\midrule
ML-100K & .5390 & .4310 & .3110 \\
ML-1M   & .6220 & .5343 & .4133 \\
\bottomrule
\end{tabular}
\caption{Mean Arm-B ValScore by candidate-pool size.}
\label{tab:pool_size}
\end{table}

Mean ValScore decreases as the candidate pool grows. Thus, $\tau=0.4$ is an operating threshold that depends on the candidate-pool construction and should be recalibrated when the verifier configuration changes.

\section{Appendix: Classical kNN Sensitivity}
\label{app:knn}

User-kNN and Item-kNN use cosine similarity over the explicit 1--5 rating matrix. Each evaluation user's row ends at the 10-item query point, excluding evaluation targets and later interactions. We report all tested $k$ settings for sensitivity analysis.

\begin{table}[h]
\centering
\small
\setlength{\tabcolsep}{4pt}
\begin{tabular}{llrrrr}
\toprule
\textbf{Data} & \textbf{Method} & $k$ & \textbf{H@1} & \textbf{H@5} & \textbf{H@10} \\
\midrule
\multirow{5}{*}{ML-100K}
& User-kNN & 20 & 4.86 & 20.28 & 38.91 \\
& User-kNN & 40 & 3.32 & 14.71 & 29.42 \\
& User-kNN & 80 & 1.54 & 10.79 & 22.89 \\
& Item-kNN & 5  & 1.30 & 4.98 & 9.25 \\
& Item-kNN & 10 & .83  & 4.51 & 8.42 \\
\midrule
\multirow{5}{*}{ML-1M}
& User-kNN & 20 & 1.73 & 10.60 & 21.40 \\
& User-kNN & 40 & 1.20 & 7.60 & 14.93 \\
& User-kNN & 80 & .67  & 4.00 & 8.13 \\
& Item-kNN & 5  & 1.33 & 4.60 & 7.93 \\
& Item-kNN & 10 & .33  & .73  & 1.13 \\
\bottomrule
\end{tabular}
\caption{Full-pool same-candidate classical collaborative baselines (\%).}
\label{tab:knn_full}
\end{table}

Classical kNN performance varies by dataset and $k$: User-kNN is relatively strong on ML-100K, whereas its effect is more limited on ML-1M.

\section{Appendix: Logged Qualitative Cases}
\label{app:cases}

The following examples are selected from actual evaluation logs and include sample indices. They illustrate the behavior of the verifier and regeneration and are not intended to represent average performance.

\paragraph{High-recoverability case --- ML-100K, sample 70.}
The history includes \emph{Searching for Bobby Fischer}, \emph{Casablanca}, \emph{Field of Dreams}, \emph{Fargo}, and \emph{The Shawshank Redemption}. The generated preference note captures emotionally resonant and character-driven narratives, moral depth, redemption, and restrained performances, with ValScore 0.9. Stage-1 and Forward-Look both predict \emph{The Remains of the Day}, and Hit@10 remains $1\rightarrow1$.

\paragraph{Successful regeneration --- ML-100K, sample 204.}
The history includes \emph{The English Patient}, \emph{Tomorrow Never Dies}, \emph{Contact}, \emph{Air Force One}, and \emph{Liar Liar}. The initial preference has ValScore 0.3. After regeneration, characteristics such as a drama--suspense mixture, high-stakes personal crisis, and interpersonal tension become more specific, increasing ValScore to 0.6. The prediction changes from \emph{Sophie's Choice} to \emph{As Good as It Gets}, and Hit@10 improves from $0\rightarrow1$.

\paragraph{Counterexample --- ML-100K, sample 529.}
The history includes \emph{Scream}, \emph{Evita}, \emph{Lost Highway}, \emph{Crash}, \emph{187}, and \emph{Volcano}. After regeneration, features such as psychological thriller, moral ambiguity, and atmospheric tone become more specific, increasing ValScore from $0.3\rightarrow0.4$. The prediction changes from \emph{Dead Man} to \emph{Chinatown}, while Hit@10 decreases from $1\rightarrow0$.

\section{Appendix: ValScore and Recommendation Correctness}
\label{app:valscore_assoc}

To examine the association between ValScore and Stage-2 recommendation correctness, we report Hit@10 by initial ValScore bucket and the sample-level Pearson correlation.

\begin{table}[h]
\centering
\small
\setlength{\tabcolsep}{4pt}
\begin{tabular}{lrr}
\toprule
\textbf{ValScore bucket} & \textbf{ML-100K Hit@10} & \textbf{ML-1M Hit@10} \\
\midrule
LOW [0,.4)   & 20.69 ($n=29$)  & 24.14 ($n=87$) \\
MID [.4,.5)  & 26.19 ($n=84$)  & 25.52 ($n=145$) \\
HIGH [.5,.7) & 24.58 ($n=301$) & 35.43 ($n=525$) \\
PASS [.7,1]  & 30.77 ($n=429$) & 40.92 ($n=743$) \\
Pearson $r$ & .044 $[-.022,.110]$ & .115 $[.065,.165]$ \\
\bottomrule
\end{tabular}
\end{table}

The sample-level Pearson correlation between ValScore and Stage-2 correctness is $r=.044$ on ML-100K (95\% CI $[-.022,.110]$) and $r=.115$ on ML-1M (95\% CI $[.065,.165]$).

\section{Appendix: Timestamp-Safe Diagnostic}
\label{app:timestamp}

In the strict timestamp-safe subset, Forward-Look improves Hit@10 over Stage-1+Regeneration by $+5.14$\,pp on ML-100K ($n=564$, 95\% CI $[+1.06,+9.22]$). On ML-1M, the gain is $+0.67$\,pp ($n=1{,}487$, 95\% CI $[-1.75,+3.09]$).

\section{Appendix: Popularity-Only Baseline Analysis}
\label{app:popularity}

\begin{table}[t]
\centering
\small
\setlength{\tabcolsep}{3pt}
\begin{tabularx}{\columnwidth}{@{}>{\raggedright\arraybackslash}Xrr@{}}
\toprule
\textbf{Decoder} & \textbf{ML-100K} & \textbf{ML-1M} \\
\midrule
Global popularity & 32.38 & 30.80 \\
Query-time-safe popularity & 37.13 & 30.13 \\
Forward-Look retrieval-only & 42.70 & 45.40 \\
Timestamp-safe Forward-Look retrieval-only & 40.69 & 42.87 \\
\bottomrule
\end{tabularx}
\caption{Hit@10 (\%) for popularity and Forward-Look retrieval-only baselines.}
\label{tab:popularity}
\end{table}

Forward-Look retrieval-only achieves higher Hit@10 than both global popularity and query-time-safe popularity baselines on both datasets.

\section{Appendix: Prompt Templates}
\label{app:prompts}

This appendix provides the full system / user prompts used at every
LLM call in the pipeline. \texttt{\{...\}} denotes runtime-filled
placeholders.

\paragraph{Stage-1 semantic preference generation
(\S\ref{sec:stage1}).}
\begin{quote}
\small
\texttt{\textbf{system}: You are a helpful assistant that analyzes a
user's movie preferences based on their watching history. Write a
concise preference description in about 60--100 words.}

\texttt{CRITICAL CONSTRAINT --- do NOT name or quote any specific
movie title from the user's history. Describe taste only in terms of:
genre / sub-genre, mood and emotional tone, thematic content, era or
stylistic period, narrative style. Mentioning a specific title from
the history will be considered a failure.}

\texttt{\textbf{user}: The user's previous order of watching movies
and their ratings were: \{history\_str\}. Please analyze his
preference characteristics based on the previous behavior.}
\end{quote}

\paragraph{Stage-1 semantic preference --- Electronics adaptation
(\S\ref{sec:elec}).}
The Electronics prompt replaces the five thematic axes with five
spec-centric axes; the preamble and constraint structure are
otherwise identical.
\begin{quote}
\small
\texttt{\textbf{system}: You are a helpful assistant that analyzes a
user's product preferences based on their purchase history. Write a
concise preference description in about 60--100 words.}

\texttt{CRITICAL CONSTRAINT --- do NOT name or quote any specific
product title from the user's history. Describe the user's
preference along five concrete technical axes: (1) technical
specifications (resolution / capacity / speed / wattage /
connectivity), (2) industry standards and compatibility (USB-C,
HDMI, IEEE, etc.), (3) form factor (size, weight, portability), (4)
brand family or product line, and (5) quality tier. Be precise and
concrete about the technical characteristics that unify their
purchases. Avoid generic adjectives.}

\texttt{\textbf{user}: The user's previous order of purchasing
products and their ratings were: \{history\_str\}. Please analyze
his preference characteristics based on the previous behavior.}
\end{quote}

\paragraph{Validator: cycle-consistency verification
(\S\ref{sec:pv}).}
\begin{quote}
\small
\texttt{\textbf{system}: You are a movie analyst. Given a user
preference description, identify which movies from a list the user
has likely watched.}

\texttt{\textbf{user}: A user has the following preference
characteristics:}\\
\texttt{\{preference\}}\\
\texttt{From the movies listed below, select the ones this user has
most likely already watched (based on the preference description).
Respond with ONLY a comma-separated list of movie titles, nothing
else.}\\
\texttt{Movies:}\\
\texttt{- \{candidate\_1\}}\\
\texttt{- \{candidate\_2\}}\\
\texttt{...}\\
\texttt{- \{candidate\_50\}}
\end{quote}
The candidate set comprises the user's true history items ($|H_u| = 10$) plus 40 random negatives, shuffled. More precisely, the original sampler uses 40 distinct catalog-wide unobserved distractors, does not restrict query-time availability, and does not explicitly remove later evaluation targets. Sensitivity to this construction is analyzed in Appendix~\ref{app:negative}.

\paragraph{Stage-2 Ranker (with Forward-Look hint,
\S\ref{sec:forward}).}
\begin{quote}
\small
\texttt{\textbf{system}: You are a movie recommendation assistant.
Use (1) the user's history, (2) their preference description, AND
(3) `next-watch hints' showing what other users with similar viewing
patterns watched NEXT --- these are candidates from a
collaborative-filtering signal. Pay attention to them. Respond with
ONLY the title of one movie from the candidate list.}

\texttt{\textbf{user}: The user's previous order of watching movies
and their ratings were: \{history\_str\}}\\
\texttt{His Personal preferences: \{preference\}}\\
\texttt{Users with similar viewing patterns next watched:}\\
\texttt{\hphantom{xx}(genre) \{future\_titles\_fragment\_1\}}\\
\texttt{\hphantom{xx}(genre) \{future\_titles\_fragment\_2\}}\\
\texttt{\hphantom{xx}\ldots}\\
\texttt{\hphantom{xx}(genre) \{future\_titles\_fragment\_k\}}\\
\texttt{The movies available for selection include: \{candidate\_1\} ;
\{candidate\_2\} ; \ldots ; \{candidate\_100\} .}\\
\texttt{Based on his viewing history and preferences, please
recommend a movie for him.}
\end{quote}

\paragraph{Stage-2 Ranker (without Forward-Look hint --- the A/C
ablation conditions).}
The hint block is removed and only the system prompt is replaced by:
\begin{quote}
\small
\texttt{\textbf{system}: You are a movie recommendation assistant.
Given a user's history, their preference description, and a
candidate list, respond with ONLY the title of the single best movie
to recommend from the candidate list.}
\end{quote}

\paragraph{Conditional Regen (\S\ref{sec:regen}).}
For regeneration, the same Stage-1 prompt is re-called at sampling
temperature $T{=}0.5$ up to $m{=}2$ times. Each attempt is scored by
the Validator prompt above, and the highest-scoring attempt becomes
$P_u^{'}$.

\end{document}